%% file: main.tex
\documentclass[format=acmsmall, review=false]{acmart}

\pdfoutput=1 

\renewcommand\footnotetextcopyrightpermission[1]{} 

\setcopyright{none}

\usepackage{graphicx}
\usepackage[caption=false]{subfig}
\usepackage{url}
\usepackage{xcolor}
\usepackage{enumitem}
\usepackage{layouts} 
\usepackage{xifthen} 
\usepackage{dsfont} 
\usepackage[normalem]{ulem} 
\usepackage{csquotes} 
\usepackage{amssymb}
\usepackage{algorithm}
\usepackage{algpseudocode}
\usepackage[capitalise,noabbrev]{cleveref} 

 \theoremstyle{acmplain}
 
\input{comands.tex}

\begin{document}

\fancyfoot{}

\title[Benchmarking Unsupervised Outlier Detection with Realistic Synthetic Data]{Benchmarking Unsupervised Outlier Detection\\ with Realistic Synthetic Data} 

\author{Georg Steinbuss}
\email{georg.steinbuss@kit.edu}
\orcid{0000-0002-2051-3394}
\affiliation{%
	\institution{Karlsruhe Institute of Technology (KIT)}
	\city{Karlsruhe}
	\country{Germany}
}
\author{Klemens B\"{o}hm}
\email{klemens.boehm@kit.edu}
\affiliation{%
	\institution{Karlsruhe Institute of Technology (KIT)}
	\city{Karlsruhe}
	\country{Germany}
}

\begin{abstract}
	Benchmarking unsupervised outlier detection is difficult.
	Outliers are rare, and existing benchmark data contains outliers with various and unknown characteristics.
	Fully synthetic data usually consists of outliers and regular instance with clear characteristics and thus allows for a more meaningful evaluation of detection methods in principle.
	Nonetheless, there have only been few attempts to include synthetic data in benchmarks for outlier detection.
	This might be due to the imprecise notion of outliers or to the difficulty to arrive at a good coverage of different domains with synthetic data.
	In this work we propose a generic process for the generation of data sets for such benchmarking.
	The core idea is to reconstruct regular instances from existing real-world benchmark data while generating outliers so that they exhibit insightful characteristics. 
	This allows both for a good coverage of domains and for helpful interpretations of results.
	We also describe three instantiations of the generic process that generate outliers with specific characteristics, like local outliers.
	A benchmark with state-of-the-art detection methods confirms that our generic process is indeed practical.
\end{abstract}



\maketitle
\thispagestyle{empty}

\section{Introduction}

Unsupervised outlier detection aims at finding data instances in an unlabeled data set that deviate from most other instances. 
Such outliers tend to be rare and usually exhibit unknown characteristics. 
This renders the evaluation of detection performance and hence comparisons of unsupervised outlier-detection methods difficult.
Researchers tend to use benchmark data sets with labeled instances to this end \citep{domingues_comparative_2018,goldstein_comparative_2016,campos_evaluation_2016,emmott_systematic_2013,emmott_meta-analysis_2015}.
Since the data sets often are from some real-world classification problem, identifying the characteristics of outliers and seeing how these characteristics compare to outliers from another data set can be very difficult. 
Often one can only approximate the process generating outliers or regular instances. 
See \cref{exa:motivation}.

\begin{figure}[ht]
	\centering
	\includegraphics[width=1\linewidth]{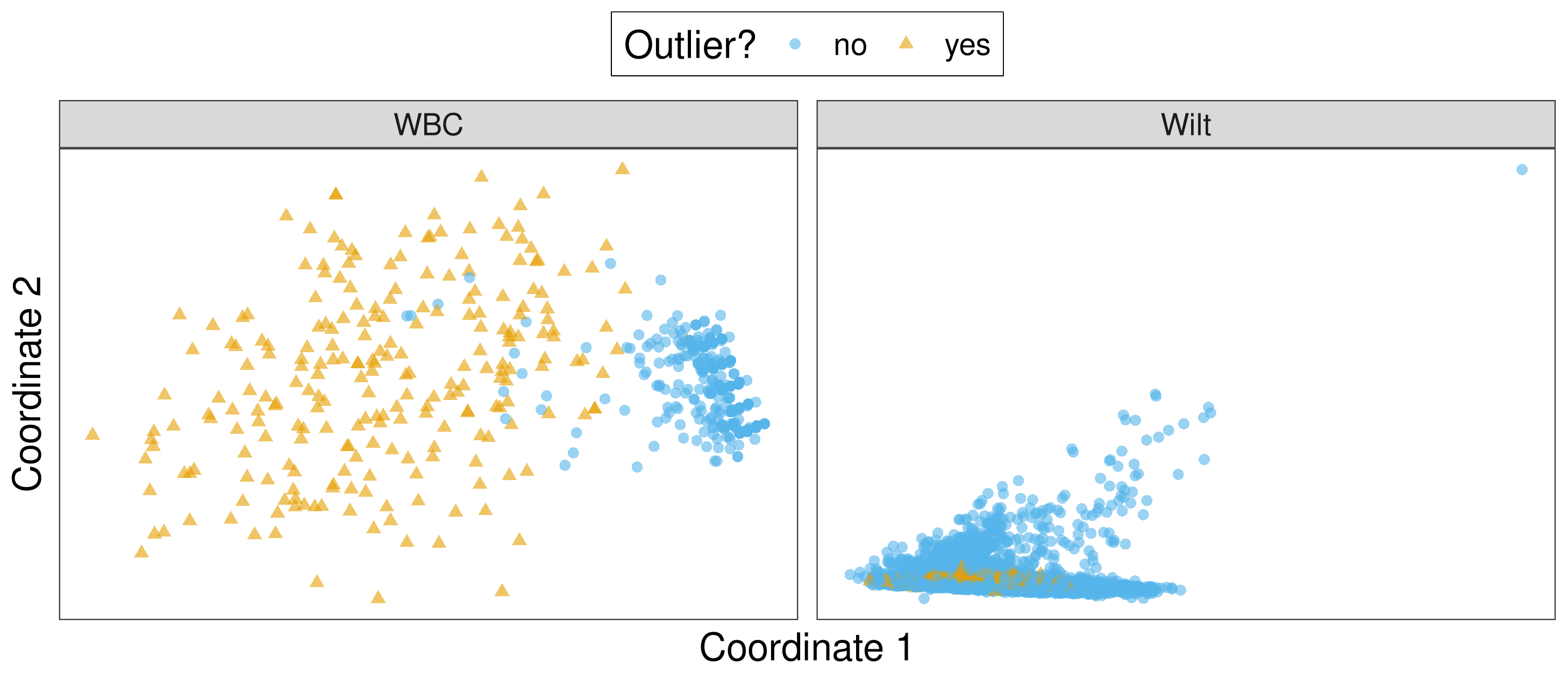}
	\caption{Result from multidimensional scaling.}
	\Description{Two exemplary data sets that feature outliers with almost opposite characteristics.}
	\label{fig:heartdisease-motivation}
\end{figure}

\begin{example}\label{exa:motivation}
	The \emph{WBC} \citep{campos_evaluation_2016} data set comprises instances on benign and malignant ($=:$ outliers) cancer. 
	The \emph{Wilt} \citep{campos_evaluation_2016} data set holds satellite images of diseased trees ($=:$ outliers) and other land cover. 
	\cref{fig:heartdisease-motivation} displays two-dimensional representatives of both data sets obtained with multidimensional scaling \citep{hastie_elements_2017}. 
	This scaling aims at keeping the pairwise distances from the original data for the lower dimensional representatives.
	The figure illustrates that the outliers from both data sets feature very different characteristics in terms of their pairwise distance. 
	The outliers from WBC are rather distant from each other and from the regular instances in particular. 
	In the Wilt data set, outliers are very close to each other and to regular instances.
\end{example}

There exist approaches that generate fully synthetic data, to evaluate that detection performance \citep{iglesias_mdcgen:_2019,domingues_comparative_2018,emmott_meta-analysis_2015,pei_synthetic_2006}. 
Fully synthetic data can contain outliers that exhibit much clearer and more consistent characteristics than the real outliers in \cref{exa:motivation}. 
However, only two generation approaches for synthetic data specific to outlier detection 
have been used in current benchmarks \citep{domingues_comparative_2018,emmott_meta-analysis_2015}. 
Both approaches are of a similar and quite simplistic nature: Distributions of outliers and regular instances are simple; for instance, \citet{emmott_meta-analysis_2015} propose a single Gaussian for regular instances and the uniform distribution for outliers. 
These approaches do not yield any insights regarding detection performance with outliers with other characteristics or with a more complex distribution of regular instances. 
In this work we propose a benchmark with the explicit objective of synthetic data being realistic and featuring outliers with insightful characteristics. 
We envision a benchmark that, by fulfilling this objective, allows for interpretations beyond pure detection accuracy. 
An example is how well certain detection methods handle outliers that only deviate from the attribute dependency exhibited by regular instances, but not from the one-dimensional distribution of values in each individual attribute. 
We refer to synthetic outliers with such specific characteristics as \emph{characterizable}.

\subsection{Challenges}

A major issue is the ambiguity of the notion of outliers. 
While there exist different intuitions regarding this, there is no general and precise definition \citep{zimek_there_2018}. 
This renders generating outliers a very difficult task. 
A common solution is to generate outliers from a uniform distribution across the instance space \citep{domingues_comparative_2018,emmott_meta-analysis_2015,melnykov_mixsim_2012,maitra_simulating_2010,qiu_generation_2006,pei_synthetic_2006}. 
However, samples from one uniform distribution mostly are global outliers, i.e., outliers far from any regular instance. 
But detection performance with other types of outliers may be of even greater interest. 
One such type of outliers are the dependency outliers mentioned earlier.  
Local outliers \citep{breunig_lof:_2000} are another outlier type of interest. 
These are outliers that deviate within their local neighborhood, for example the closest cluster.
The type of outliers can also depend on the characteristics of regular instances:
If one wants to generate, say, local outliers, one needs to characterize the local neighborhood, which mostly consists of regular instances.
If one does not really know the characteristics of regular instances in this local region (e.g., since they are estimated), synthetic outliers might not really be local.
Thus, simply adding some generated outliers to real regular instances is not sufficient to generate outliers of specific types.

Another challenge is coverage. 
We hypothesize that a single detection method does not outperform other methods in every data domain.
Thus, a proper benchmark for detection methods must encompass a variety of data domains.
Existing benchmarks \citep{domingues_comparative_2018,goldstein_comparative_2016,campos_evaluation_2016,emmott_systematic_2013,emmott_meta-analysis_2015} approach this by using many data sets from different domains. 
With synthetic data, coverage of several domains is challenging. Data generators for a certain domain are usually handcrafted and hence difficult to compare to each other. 
To illustrate, \citet{barse_synthesizing_2003} generate data from the fraud detection domain by simulating users, \citet{downs_plant-wide_1993} do so for data from a plant by modeling chemical processes. 
Synthetic data generators that are domain agnostic like MDCGen \citep{iglesias_mdcgen:_2019} need to be parameterized. 
For example, the number of clusters or their form need to be specified. 
To our knowledge, there currently is no widely accepted way to choose these parameters so that the resulting data sets cover different domains sufficiently.

\subsection{Our Approach}

The idea central to our work is to use a real-world data set as a basis for generating a synthetic one. 
Both regular instances and outliers are generated, but differently:
Synthetic regular instances follow a model of the existing real regular instances.
Outliers are in line with a characterizable deviation from this model.
In what follows, we refer to these steps as \enquote{process}. 
Put differently, this article proposes a process with this fundamental design, with various ramifications: 
The regular instances are realistic, and repeating the process for different real-world data sets also gives way to good coverage, i.e., consideration of different domains.
On the other side, generating outliers similar to the ones labeled as such in real-world data would not improve the insights obtainable from a respective benchmark by much: 
The generated outliers would be too diverse to allow for meaningful conclusions, since they would exhibit the characteristics of the real outliers (cf.\ \cref{exa:motivation}).
This is why we propose to generate outliers as a characterizable deviation from the model of regular instances. 
Creating this deviation usually needs parameters.
But in contrast to parameters of many existing data generators, like the number of clusters of the regular instances, these parameters directly specify the characteristics of the generated outliers:
To illustrate, think of the distance of local outliers to their local neighborhood. 
Thus, the parameters of our process can be useful to control the characteristics of outliers.

To have good interpretability, we demand that for any generated instance --- be it outlying, be it regular --- we have access to its probability density. 
This allows for the introduction of several ideal outlier scorings, as we call them.
One is for example similar to the Bayes error rate \citep{tumer_estimating_1996} from supervised classification tasks. 
This rate gives insight into the prediction error that is unavoidable.
In our case this error comes from generated outliers that are indistinguishable from regular instances. 
Statistical distributions (like Gaussian Mixtures) naturally give access to the probability density, are powerful, i.e., allow to model diverse data distributions, and allow for the generation of characterizable outliers.
The body of the article will provide further reasons why there is this focus on statistical distributions.

\subsection{Contributions}
\label{sec:contribution}

Our first contribution is formalizing the generic process sketched so far.
In essence, the process relies on real-world data and on models to generate similar regular data as well as characterizable outliers. 
Our second contribution is a proof-of-concept instantiation of our benchmark: We show that the process does allow for conclusions on the performance of outlier-detection methods with different characterizations of outliers, i.e., with different characterizable deviations from regular instances. 
The three specific instantiations used here are as follows: one for local outliers, one for dependency outliers, and one for global outliers.

While the number of outlier characteristics is infinite (cf.\ \cref{sec:infinite_types_outliers}), and while identifying the best detection method for a full range of domains and outlier characteristics is beyond the scope of one article, for the specific case inspected here, we were already able to find that dependency outliers are best found by a very local version of the Local Outlier Factor (LOF) \citep{breunig_lof:_2000}.
Our third contribution is to validate the realness of our synthetic data, by means of experiments.
They confirm that the instantiations just mentioned do result in synthetic data close to the respective real-world data set.

\subsection{Article Structure}

\cref{sec:relwork} reviews related work. 
In \cref{sec:process} we propose a general process to generate realistic synthetic data. 
In \cref{sec:instantiations} we introduce three instantiations of this process for our benchmark. 
In \cref{sec:design-of-experiments} we describe the design of our experiments.
In \cref{sec:realness} and \cref{sec:benchmark} we discuss our results, and \cref{sec:conclusion} concludes.

\section{Related Work}
\label{sec:relwork}

Our related work has three parts. We first review benchmarks for unsupervised outlier detection. 
The second part deals with synthetic data generation in general and the last one with the generation of outliers.

\subsection{Benchmarks for Outlier Detection}

Several benchmarks for unsupervised outlier detection exist \citep{domingues_comparative_2018,goldstein_comparative_2016,campos_evaluation_2016,emmott_systematic_2013,emmott_meta-analysis_2015}. 
They all use real-world data, usually from classification. One class that is semantically meaningful (e.g., the patients with a disease from \cref{exa:motivation}) is defined as the outlier class. 
It is usually downsampled since outliers should be rare. 
This does not solve the issue from \cref{exa:motivation}. 
\cite{emmott_systematic_2013,emmott_meta-analysis_2015} introduce problem dimensions for a more systematic benchmark with real-world data. 
These dimensions are point difficulty, relative frequency, semantic variation and feature relevance. 
The articles also propose approaches to measure these dimensions in real-world data. 
By sampling specific instances they vary these dimensions within the data and use these more controlled versions to benchmark detection methods.
This does improve the situation from \cref{exa:motivation}, but heavily depends on the approaches used to quantify the proposed dimensions. 
For example, semantic variation is measured by comparing variance estimates for the outliers and regular instances. 
Clearly this is a very crude approximation. 
Think of the outliers as instances of two tight clusters that are far apart. 
In addition, with this approach outliers are not easily characterizable in the sense introduced earlier. 
So the interpretation of detection performance in this regard is not feasible.

\subsection{Synthetic Data}
\label{sec:related_synthetic_data}

There is a lot of literature on synthetic data, but here we focus on outlier detection. For broader reviews see  \citep{zimmermann_method_2019,cliquet_realistic_2019,frasch_bayes-true_2011,steinley_oclus:_2005}. 
First we review domain specific data generators, then generators developed without a specific domain in mind. 
Lastly we review generation approaches that use real data as a basis, as done here.

\subsubsection{Domain Specific}
\label{sec:synth-domain-specific}

Many synthetic data sets are specific to a certain domain. 
I.e., their generation process is designed with a certain application in mind, like chemical processes \citep{downs_plant-wide_1993}, fraud detection \citep{barse_synthesizing_2003} or application scoring \citep{kennedy_framework_2011}. 
Such synthetic data is already used in benchmarks for unsupervised outlier detection. 
For example, in the benchmark by \citet{campos_evaluation_2016} the data set Waveform\footnote{\url{http://archive.ics.uci.edu/ml/datasets/waveform+database+generator+(version+2)}} is synthetic. 
Such data can be useful when comparing outlier detection approaches. 
However, domain specific synthetic data tends to be generated not for outlier detection in particular.
For instance, the Waveform data is for classification. 
Hence, if it really is a good benchmark for outlier detection remains  questionable. 
Even if it inherits some class that is semantically meaningful as outliers, like \citep{downs_plant-wide_1993,barse_synthesizing_2003,kennedy_framework_2011}, it is very resource demanding to generate such data. 
One reason is that a domain expert must be involved in its design. 
Novel frameworks for crafting realistic data like the one introduced by \citet{mannino_is_2019} might reduce the effort by utilizing smart visualizations and useful suggestions for attribute values, distributions and dependencies among them. 
However, an expert still must be involved. 
Hence, a good coverage of a vast amount of different data domains is difficult to achieve. 

\subsubsection{Domain Agnostic}
\label{sec:synth-domain-agnostic}

There exist many synthetic data generators not situated in a certain domain, most of them for clustering \citep{iglesias_mdcgen:_2019,hutchison_n-spheres_2013,melnykov_mixsim_2012,maitra_simulating_2010,qiu_generation_2006,pei_synthetic_2006,steinley_oclus:_2005,waller_method_1999,milligan_algorithm_1985} or classification \citep{frasch_bayes-true_2011,rachkovskij_datagen:_1998}. 
A major difference between the generators for clustered data is the control of overlap \citep{hutchison_n-spheres_2013,melnykov_mixsim_2012,maitra_simulating_2010,qiu_generation_2006,steinley_oclus:_2005,milligan_algorithm_1985}. How to parameterize the generators so that they cover diverse domains remains an open question. 
Next, the probability densities are not always available. 
However, the way we generate data is not far from ideas behind these generators. 
Some use for example Gaussian mixtures \citep{hutchison_n-spheres_2013,melnykov_mixsim_2012,frasch_bayes-true_2011,maitra_simulating_2010,milligan_algorithm_1985} as we do for some instantiations as well.

\subsubsection{From Real Data}
 \label{sec:related_synth_from_real}

Synthetic reconstruction aims at generating data that matches given real data. 
The term \enquote{synthetic reconstruction} is known from survey data \citep{wan_sync:_2019}
but the principle is common in other disciplines as well \citep{mannino_is_2019,sun_learning_2018,benczur_statistical_2018,albuquerque_synthetic_2011,waller_method_1999}. 
Often synthetic reconstruction is needed due to a limited amount of data or due to privacy issues with the original data \citep{cliquet_realistic_2019}. 
Some data generators for purely synthetic data mentioned earlier allow for adaptation to real-world data. 
The frameworks of \citet{albuquerque_synthetic_2011} and \citet{iglesias_mdcgen:_2019} allow for user-defined distributions. 
The example of \cite{waller_method_1999} where they adopt parameters of their generation to real-world data is similar. 
Fitting statistical distributions to obtain realistic synthetic data is not uncommon either \citep{sun_learning_2018,benczur_statistical_2018,rogers_using_2003}. 
For example, \citet{rogers_using_2003} use Gaussian mixtures to model protein spots in images of electrophoresis gel. 
Synthetic data is then generated from this model and used for evaluation purposes. 
Approaches like Generative Adversarial Networks \citep{goodfellow_generative_2014} recently received much attention for their astonishing capabilities in generating realistic synthetic data. 
Here we concentrate on approaches like \citep{sun_learning_2018,benczur_statistical_2018,rogers_using_2003} that fit statistical distributions to real-world data and draw samples from these models for generating data. The reasons for this focus are manifold, and we will cover them in \cref{sec:fit_regular}.

\subsection{Generating Outliers}

Two benchmarks reviewed earlier \citep{domingues_comparative_2018,emmott_meta-analysis_2015} feature synthetic data with outliers.
 \citet{emmott_meta-analysis_2015} generate data with regular instances following a simple multivariate Gaussian distribution, while \citet{domingues_comparative_2018} generate regular instances from two separate Students' T distributions.
 In both cases, outliers come from a uniform distribution surrounding the regular instances. 
\citet{emmott_meta-analysis_2015} already state themselves that such simple data is not necessarily helpful.

Generators for clustering introduced in \cref{sec:synth-domain-agnostic} also feature the generation of outliers, usually from a uniform distribution within the instance space \citep{iglesias_mdcgen:_2019,melnykov_mixsim_2012,maitra_simulating_2010,qiu_generation_2006,pei_synthetic_2006}. 
\citet{iglesias_mdcgen:_2019} generate outliers not by sampling from a uniform distribution, but by using the intersections of a grid. 
While we still deem this close to uniformity, it lets them generate subspace outliers. 
\citet{pei_synthetic_2006} generate outliers that follow pre-specified patterns. 
In their experiments they use lines to this end. 
\citet{milligan_algorithm_1985} proposes to generate outliers by sampling form a Gaussian with increased variance in comparison to the Gaussian regular instances are sampled from. 
This is interesting here since it yields local outliers. 
Thus, we will use it for one instantiation. 

There also exist approaches that generate outliers based on real regular instances.
In this regard, the uniform distribution is quite common \citep{hastie_elements_2017,steinwart_classication_2005,steinbuss_hiding_2017,theiler_resampling_2003,tax_uniform_2001}, but many other approaches exist as well:
For example, \citet{vernekar_out--distribution_2019} propose to generate outliers close to the regular instances using an autoencoder, or \citet{tax_uniform_2001} propose to generate outliers uniformly from a hypersphere surrounding the regular instances.
One issue with these approaches is that the characteristics of the resulting outliers often are not easy to interpret. 
For example, the outliers generated with the approach proposed by \citet{wang_hyperparameter_2018} surround regular instances closely and, hence, follow the boundary of regular instances, which often is hard to describe.
While the similarity to the boundary of regular instances
seems useful to estimate parameters of a one-class classifier, it is unknown how well the generated instances represent outliers in an unsupervised scenario. 
Another issue is that there usually is no model of the real regular instances. 
Hence, their density and often also the density of the generated outliers remain unknown. 
The computation of ideal scorings to compare with --- like in our approach --- is thus not possible. 
However, some approaches for the generation of outliers based on real regular instances are very close to one of our instantiations. 
Outliers generated by sampling instance values independently \citep{hastie_elements_2017,theiler_resampling_2003} do not follow the dependency structure of the data attributes but remain within the distribution of each attribute for itself. 
Outliers from one of our instantiations (\cref{sec:vine}) behave like this as well.

\section{The Generic Process}
\label{sec:process}

In this section we formalize the generic process we propose for realistic synthetic benchmarks for unsupervised outlier detection. 
First we introduce some notions, then our requirements regarding instantiations of the generative process. 
We then introduce our ideal scorings and provide an overview of the process.

\subsection{Fundamentals}

A real-world data set $\realData$ is a set of $\nReal$ instances $\instance$. 
Each $\instance$ is a vector from a $\nDim$-dimensional subset of $\R^\nDim$. 
The same holds for a synthetic reconstruction of $\realData$ denoted by $\synthData$. 

A generative model $\model \in \Omega$ is a description of a set of instances that allows for the generation of synthetic instances. 
$\Omega$ is the set of all possible models, used here for purely notational purposes. 
A generative model in our context is associated with the four functions
\begin{equation*}
    \fitNoArg \colon \R^{\nReal \times \nDim} \to \Omega, \
    \genNoArg \colon \Omega \to \R^{\nReal \times \nDim}, \ 
    \densNoArg \colon \R^{\nReal \times \nDim} \times \Omega \to \R^{\nReal}, \
    \modifyNoArg \colon \Omega \to \Omega .
\end{equation*}
$\fitNoArg$ creates the generative model from a set of instances. 
$\genNoArg$ generates instances from the generative model.  
$\densNoArg$ returns the density of a set of instances in terms of a model. 
$\modifyNoArg$ modifies a generative model so that the resulting model can generate outliers that are characterizable.

An unsupervised outlier detection method is denoted as function $\dectNoArg \colon \R^{\nReal \! \times \! \nDim} \to \R^{\nReal}$.
It outputs a score, given a set of instances. 
The scores must have a meaningful ordering in terms of outliers. 
For example, with the LOF a high score indicates outliers.

\subsection{Requirements on Generation}
\label{sec:fit_regular}

In \cref{sec:contribution} we have argued that fitting the labeled regular instances \emph{and} the outliers from real-world data is not very insightful. 
Hence, in this benchmark we reconstruct only the regular instances by fitting a generative model to them.  
Outliers are then generated based on a \emph{characterizable} deviation from this model.

According to \cref{sec:related_synth_from_real}, there already exist approaches for the synthetic reconstruction of a data set (basically $\fitNoArg$ and $\genNoArg$). 
To decide on a suitable one for our process we gather requirements that it must fulfill.
\begin{enumerate}[label=\underline{R\arabic*:},itemsep=0pt]
	\item Be applicable to many kinds of real-world data. \label{itm:widely_applicabble}
	\item Generate realistic synthetic data.
	\item Feature a generation that is comprehensible.
	\item Give access to the density of generated instances.
	\item Allow for generating characterizable outliers.
\end{enumerate}

R1 and R2 are important to achieve a good coverage of different data domains and R3 to R5 for a good interpretation of the results from a benchmark. 
Except for R4 the requirements do not allow for a rigid definition. 
Terms like \emph{many kinds of real-world data}, \emph{realistic synthetic data} or \emph{comprehensibility} are subjective. 
To simplify the search for a suitable approach in terms of R2, \cref{sec:realness} will showcase a possible approach to assess the realness of synthetic reconstructions.

When generating data, we rely on statistical distributions.
Such distributions tend to fulfill R1--R5 well: 
Naturally they allow for access to the density of instances and thus fulfill R4.
In \cref{sec:instantiations}, we will show that the statistical distributions we use (e.g., Gaussian mixtures) also allow for characterizable outliers (R5).
We will also show, in \cref{sec:realness}, that these distributions are reasonably realistic (R2).
Next, statistical distributions are also applicable to almost any type of data, so they fulfill R1 as well.
Finally, models (abbreviated with $\model$ in what follows) that describe statistical distributions are well understood (R3).

\subsection{Ideal Scores}
\label{sec:ideal_scores}

This section introduces our so-called ideal scorings.
The wording \enquote{ideal scoring} highlights that there is no estimation uncertainty.
We propose different scorings. 
Each one captures different aspects that we deem important in the context of outlier detection, and one can regard each scoring as an ideal detection method.
For example, one scoring represents an ideal detection method that does know the distribution of regular instances and outliers.
\cref{def:ideal_scores} introduces our three ideal scorings Classify (C), Regular Density (RD) and Overall Density (OD). 

\begin{definition}[Ideal Scorings]\label{def:ideal_scores}
	Let $\densreg \in \R$ be the density of an instance with regard to the distribution of regular instances and $\densout \in \R$ the one with regard to the distribution of outliers. 
	Then the ideal scores for this instance are
	\begin{equation*}
	\text{RD} = \densreg, \
	\text{OD} = \xi \cdot \densout + (1-\xi) \cdot \densreg, \
	\text{C} = \xi \cdot\densout / (1-\xi) \cdot \densreg
	\end{equation*}
\end{definition}
RD and OD give outliers lower scores, and C assigns them higher scores. $\xi$ is the frequency of outliers.

Each scoring gives insights regarding characteristics of the generated data. 
C represents an ideal classifier that knows both probability distributions --- similarly to the idea of the Bayes error rate \citep{tumer_estimating_1996}.  
The nominator is proportional to the probability of an instance being an outlier and the denominator to the one of being regular. 
Thus, a score greater 1 is a prediction for outliers and a lower score for regular instances. 
The score gives insight regarding the possibility to distinguish between outliers and regular instances.
This is similar to the \emph{difficulty} of an outlier detection problem introduced in \citep{emmott_meta-analysis_2015,emmott_systematic_2013}. 
Since C emulates a supervised scenario (where outliers and regular instances can be characterized), detection performance achievable in an unsupervised setting might be much lower.

RD and OD are both densities. The idea is that unsupervised outlier detection is very close to density estimation.
The existence of many density based approaches for outlier detection and also the strong connection of distance based ones to density \citep{zimek_there_2018} support this.
RD is the density regarding the distribution of regular instances. 
Hence, any instance that is different form regular ones will have a low score. 
OD is the overall density of the data, i.e., regarding the outliers \emph{and} regular instances. The difference between RD and OD is that highly clustered outliers will have a low score in RD but not in OD. 
However, OD is much closer to the unsupervised setting in which it is very hard to characterize regular instance on their own. 
Hence, OD in particular can give insights into an ideal unsupervised scoring.

\subsection{Overview of the Process}

\cref{alg:framework} gives an overview of the benchmark process with a specific instantiation of the generation process for synthetic data (i.e., $\fitNoArg$, $\genNoArg$, $\dectNoArg$ and $\modifyNoArg$).
The input of the algorithm is a set of real-world data sets, multiple unsupervised detection methods and the frequency of outliers ($\xi$). 
For each data set a generative model is fitted to the regular instances, which results in $\mReg$ (line 3). 
In line 4, this model is modified to yield a model for outlier generation (i.e., $\mOut$). 
We provide further details on this modification for our instantiation in \cref{sec:instantiations}.
In lines 5 -- 7, both models are used to create the synthetic data.
The model $\mReg$ is used to generate regular instances and $\mOut$ to generate outliers.
Then the densities from both models are used to compute the ideal scorings, cf.\ \cref{def:ideal_scores}, in lines 8 -- 10. 
Afterwards each detection method is applied to the synthetic data in line 12, and its detection performance is computed in line 13. In line 14 the relationship to the ideal scorings is determined. 
For example, one can do this by computing the correlation of the scorings.

\begin{algorithm}[ht]
	\caption{Overview of our benchmark process.}
	\label{alg:framework}
	\begin{algorithmic}[1]
		\Require Set of $\realData$, set of $\dect{}$ and $\xi \in [0,1]$
		\For{every $\realData$}
			\State $\textit{Regulars} = $ Regular instances from $\realData$
			\State $\mReg= \fit{\textit{Regulars}}$
			\State $\mOut = \modify{\mReg}$
			\State $\textit{synth}_{\textit{Reg}} = (1-\xi) \cdot \nReal$ instances with $\gen{\mReg}$
			\State $\textit{synth}_{\textit{Out}} = \xi \cdot \nReal$ instances with $\gen{\mOut}$
			\State $\synthData = \textit{synthReg} \cup \textit{synthOut}$
			\State $\densreg = \dens{\synthData}{\mReg}$
			\State $\densout = \dens{\synthData}{\mOut}$
			\State Compute ideal scorings
			\For{every $\dect{}$}
				\State $\textit{Scores}_{\dectNoArg} = \dect{\synthData}$
				\State Compute detection performance
				\State  Determine relation of $\textit{Scores}_{\dectNoArg}$ and ideals
			\EndFor
		\EndFor
	\end{algorithmic}
\end{algorithm}

\section{Our Instantiations}
\label{sec:instantiations}

Here we introduce our instantiations to the generic process, first for local outliers, then for outliers in the dependency structure and lastly for global outliers.

With the statistical distributions we use for generating data, most computational resources are spent on fitting a model to real-world data (i.e., $\fit{}$ in line~3 of \cref{alg:framework}).
Thus, we will shortly discuss the computational effort of $\fit{}$ with our instantiations.
Since a comparison of the many available implementations for specific statistical distributions goes beyond the scope of this article, we confine our discussion of performance characteristics to the implementations used here.

\subsection{Local Outliers}

Local outliers are outlying "relative to their local neighborhoods [...]" \citep{breunig_lof:_2000}. 
Here we define a local neighborhood to be a cluster. 
More specifically, we follow the approach from \cite{milligan_algorithm_1985} to generate outliers in synthetic data from Gaussian Mixtures. 
$\fit{}$ returns the parameters of a Gaussian Mixture: the number of components $G \in \mathbb{N}$, the mixing proportion $\pi \in [0,1]^G$, the mean vector of each component $\mu_i \in \R^\nDim$ and the covariance matrix of each component $\Sigma_i \in \R^{\nDim \times \nDim}$. 
The model for generating outliers has the same parameters, 
but the covariance matrix is scaled with $\alpha > 1$ to generate the local outliers. 
I.e., $\widetilde{\Sigma}_i = \alpha \Sigma_i$ where $i \in 1,\dots,G$. 
Instances generated with the increased covariance matrix are still close to regular instances generated from the cluster with the same $\pi_i$. 
But they will often be outlying the regular instances in this cluster. 
So they mostly are local outliers.

Although \citet{milligan_algorithm_1985} proposes $\alpha = 9$, we use $\alpha = 5$. 
This yields outliers that are detectable, i.e., sufficiently far away from regular instances, but still close to the corresponding cluster of regular instances. 
This is important since the current characterization is the one of local outliers. 
We choose the value of $G$ from the range of $1, \dots, 9$ using the BIC \citep{hastie_elements_2017}.

We use the \href{https://cran.r-project.org/web/packages/mclust/mclust.pdf}{mclust} R package \citep{scrucca_mclust_2016} for implementing Gaussian mixtures.
To model specific cluster shapes, \href{https://cran.r-project.org/web/packages/mclust/mclust.pdf}{mclust} allows for specifying different forms of $\Sigma_i$.
In the following, we refer to these different forms as \enquote{configurations}.
For example, with the configuration named \enquote{EII}, each cluster has the same spherical shape and size; this is in line with clusters found by the well-known k-means algorithm. 
The number of values that need to be estimated differs between configurations. 
While additional parameters usually allow for modeling more complex clusters, they also might bring down the computational performance. 
In preliminary experiments of ours, we found that the \enquote{VEI} configuration seems to be a good trade off between modeling power (i.e., how well the models fit the data), stable results (i.e., that one always finds a model) and computational performance.
Regarding the overall performance of our instantiation with Gaussian mixtures, even an off-the-shelf laptop allowed for fitting configurations more complex than \enquote{VEI} to data sets of medium size (27 attributes and 768 observations) in under a second.

\subsection{Dependency Outliers}
\label{sec:vine}

Let $F_i(\cdot)$ be the cumulative distribution and $f_i(\cdot)$ the probability density function of data attribute $i$. 
If the distribution of each attribute is absolutely continuous, the full multivariate probability density function $f(\cdot)$ can be written \citep{aas_pair-copula_2009} as
\begin{equation*}
	f(x_1, \dots, x_\nDim) = f(x_1) \cdot \ldots \cdot f(x_\nDim) \ \cdot \ c(F_1^{-1}(x_1), \dots, F_\nDim^{-1}(x_\nDim)).
\end{equation*}
The copula $c(\cdot)$ provides "a way of isolating the [...]  dependency  structure." \citep{aas_pair-copula_2009}. 
We use it to generate outliers that do not follow the dependency structure which regular instances follow. 
Outliers in the dependency have recently attracted some attention (see for instance the work of  \citet{ren_fault_2017}), and there also exist proposals for generating outliers in this spirit \citep{hastie_elements_2017,theiler_resampling_2003}. 
A powerful variant of modeling the copula is the vine \citep{aas_pair-copula_2009}. 
In principle, the multivariate distribution is decomposed into multiple bi-variate building blocks. 
Each bi-variate distribution can be rather simple but their combination allows modeling complex dependency patterns.

Here, $\fit{}$ returns distribution estimates for each attribute and a vine copula. 
$\modify{}$ returns the distribution estimates for the attributes without any modification of the estimates. 
However, the vine copula is set to complete independence. 
Hence, generated outliers will not follow any dependency. 
To estimate the distributions of each attribute we use Kernel Density Estimation (KDE) \citep{hastie_elements_2017}. 
To select the distribution family\footnote{We choose between all families provided by the \href{https://cran.r-project.org/web/packages/rvinecopulib/index.html}{rvinecopulib} R package.} for each bi-variate copula in the vine we make again use of the BIC \citep{hastie_elements_2017}.

The complexity of vines is quadratic in the number of dimensions ($\nDim$) \citep{brechmann_truncation_2015}.
So fitting it to data with many dimensions is computationally expensive.
This effect is also noticeable with the computations necessary for our benchmark. 
With the implementation we use (the \href{https://cran.r-project.org/web/packages/rvinecopulib/index.html}{rvinecopulib} R package.), the effect is so significant that we had to exclude some high-dimensional data sets.

\subsection{Global Outliers}

Generating uniform outliers has been described many times \citep{iglesias_mdcgen:_2019,melnykov_mixsim_2012,maitra_simulating_2010,qiu_generation_2006,pei_synthetic_2006}. 
Since they scatter across the whole instance space, most samples from a uniform distribution, when outlying, are \emph{global outliers}.

The $\fit{}$ function here has three possible forms. 
In one form it returns a uniform distribution with its bounds being the minimum and maximum of each attribute. 
Clearly this is a crude and rather unrealistic fit to the given real regular instances. 
Hence, we also allow $\fit{}$ to be the corresponding function from our previous two instantiations. 
I.e., the other two forms return a Gaussian Mixture or a vine copula fitted to the data. 
With any form of $\fit{}$, $\modify{}$ returns a uniform distribution. 
Its bounds also base on the maximum and minimum of each attribute, but the values are increased by 10\%. 
Hence, even when synthetic regular instances are from a uniform distribution, there are outliers generated.

With this instantiation, computational performance clearly depends on the form of $\fit{}$.
For Gaussian Mixtures and vine copulas, we discussed computational performance in the previous two sections.
To fit a uniform distribution, we just compute the minimum and maximum of each attribute.
This computation does not have any significant effect on the overall performance.

\section{Design of Experiments}
\label{sec:design-of-experiments}

In this section we describe our experiments. 
We first describe their objectives and then give a broad overview. 
Afterwards we introduce the real-world data sets used.

\subsection{Objectives}

We list three aims we pursue with our experiments.
\begin{enumerate}[label=\underline{A\arabic*:},itemsep=0pt]
	\item An important point with our synthetic data is that regular instances in particular should be realistic. 
	One aim of our experiments is to validate this realness.
	\item So far, we have described our process to benchmark unsupervised outlier-detection methods. 
	Another rationale behind our experiments is to showcase it, i.e., we want to show how to actually perform a benchmark with our process.
	\item With our experiments we aim at a validation of our process. 
	We study whether the different characteristics of outliers really are controllable and interpretable, for our proposed instantiations from \cref{sec:instantiations}.
\end{enumerate}
While we address A1 with a separate set of experiments, we address A2 and A3 by actually performing a benchmark with our process and comparing its results to a benchmark based on real data.

At this point, since it would exceed the scope of one publication, we do not actually do a full benchmark of detection methods. 
I.e., our current objective is not to find the best performing detection method for each domain and outlier characteristic.

\subsection{Overview}

Our experiments were coded in the \href{https://www.r-project.org/}{R language} using the \href{https://cran.r-project.org/web/packages/batchtools/batchtools.pdf}{batchtools} package \citep{lang_batchtools:_2017} for organization and parallelism\footnote{There is a copy of our code available at \url{http://ipd.kit.edu/mitarbeiter/steinbussg/synth-benchmark-code-V3.zip}.}. 
Thus, to illustrate the workflow of respective experiments we use two entities from batchtools: problems and algorithms.

A problem describes the data an algorithm is used on. 
Any problem is based on a real-world data set and might replace regular instances, outliers or both with synthetic examples. 
Thus, for each real-world data set described in \cref{sec:datasets} there is four problem types.
\begin{itemize}[itemsep=0pt]
	\item \textsc{Real}: The real-world data itself
	\item \textsc{SynthRegular}: Regular instances are synthetically reconstructed but outliers remain the ones from the real data, i.e., genuine.
	\item \textsc{SynthOutliers}: Outliers are synthetic (not necessarily reconstructing the real ones), but regular instances are genuine.
	\item \textsc{Synth}: Outliers are synthetic, and regular instances are synthetically reconstructed.
\end{itemize}
For the benchmark (A2 and A3) we make use of \textsc{Real} and \textsc{Synth}. 
The other problems are used to assess the realness of our data (A1).

An algorithm is a classifier (\textsc{Classify}) to assess the realness of our data or an outlier detection method like LOF (\textsc{Detect}) for the benchmark. 
We now describe what the two algorithms are used for and how.

\subsubsection{\textsc{Classify}} 

The idea behind the \textsc{Classify} algorithm was proposed by \cite{sun_learning_2018}, to show how well their model fits the real-world data. 
Essentially, it is tested how much worse the classifier performs when trained on synthetic instead of real data. 
Given a problem and a real-world data set, the workflow is as follows.
\begin{enumerate}[label=\underline{S\arabic*:},itemsep=0pt]
	\item Split the real-world data into 70\% training and 30\% test with unchanged class frequencies.
	\item Apply the problem on the training data. For example, with \textsc{SynthRegular} replace the regular instances with a synthetic reconstruction. The \textsc{Real} problem leaves the training data unchanged.
	\item Train a classifier to distinguish outliers and regular instances in the possibly synthetic training data.
	\item Report the performance of the classifier evaluated on the test data.
\end{enumerate}
Since the classifier is always tested on unseen and non-synthetic data, the drop in classification performance using synthetic data can serve as an indication of the realness of the synthetic data. 
If the synthetic data is close to the real data, the drop should be small. 

For the classifier we used a random forest implemented by the \href{https://cran.r-project.org/web/packages/ranger/ranger.pdf}{ranger} package and optimized its parameter using 10 repetitions of 10 fold cross validation from the caret package\footnote{For the grid of possible parameter values of the random forest we used the defaults from the caret package.}. 
Since the two classes in our data are usually imbalanced, classification performance is measured with Cohen's Kappa \citep{porwik_feature_2016}. 
It lies in [-1,1] where 1 indicates perfect agreement of prediction and ground truth, 0 stands for random guessing, and anything below 0 indicates a prediction worse than this. 

\subsubsection{\textsc{Detect}} 

The \textsc{Detect} algorithm performs our benchmark for unsupervised outlier detection methods.
\textsc{Detect} used with the \textsc{Real} problem is just a conventional benchmark. 
With \textsc{Synth}, it follows our general process given in \cref{alg:framework}. 
In both cases outliers are up to 5\% of the entire data. 
With \textsc{Real} this percentage is achieved by sampling from the outliers, like in \citep{campos_evaluation_2016}. 

In our benchmark we compare the outlier detection methods that are competitive according to existing benchmarks on real-world data \citep{domingues_comparative_2018,goldstein_comparative_2016,campos_evaluation_2016,emmott_systematic_2013,emmott_meta-analysis_2015}. 
While \citep{domingues_comparative_2018,emmott_systematic_2013,emmott_meta-analysis_2015} recommend the Isolation Forest \citep{liu_isolation_2008}, the benchmarks \citep{goldstein_comparative_2016,campos_evaluation_2016} result in recommendations for $k$NN \citep{ramaswamy_efcient_2000} and the Local Outlier Factor (LOF) \citep{breunig_lof:_2000}. 
In addition to these methods we included Kernel Density Estimation (KDE) \citep{hastie_elements_2017} since our ideal scores are strongly related to density. 

We tried using some meaningful default parameters and not vary them extensively, to keep our experiments comprehensible and to reduce their run time. 
For the Isolation Forest, we used the parameter values proposed in its original publication \citep{liu_isolation_2008}, i.e., $\psi =256$ and $t=100$. 
For the parameters minPts from LOF and $k$ from $k$NN, we used the values $5$ and $100$. 
While 5 is a very local choice, 100 is a rather global one. 
Instead of the plain $k$NN we used the weighted version w$k$NN \citep{goos_fast_2002}, which yields better density estimates \citep{biau_weighted_2011}.
For the bandwidth parameter $\gamma$ of KDE, we used the default rule of thumb from the \href{https://cran.r-project.org/web/packages/np/np.pdf}{np} package.

\subsection{Data Sets}
\label{sec:datasets}

We used most of the data sets proposed in \citep{campos_evaluation_2016}. 
This results in a total of 19 different data sets. We excluded \emph{Lymphography} and \emph{InternetAds} due to the categorical nature of their attributes: None of their attributes had more than 10 distinct values. 
The data sets \emph{Arrhythmia} with 166 attributes and \emph{SpamBase} with 53 attributes were excluded to keep the run time of our experiments at a reasonable level. 
We also kept domain-specific synthetic data sets as described in \cref{sec:synth-domain-specific}. 
\emph{Waveform} is an example. 
Such data has been engineered for a specific real-world use case, and we deem it realistic.

Similarly to \cite{campos_evaluation_2016} we only kept numeric attributes in each data set, removed duplicate values and normalized each attribute to [0,1]. 
We also excluded any attribute with less than 10 distinct values. In these cases, the attribute is most likely categorical (e.g., gender coded with \{0, 1\}) and fitting a continuous distribution yields unrealistic synthetic data. 
When we fit statistical distributions to the data, we add some noise to the attribute values according to the procedure described in \citep{nagler_generic_2017}. This should improve the fitting process.

\section{Results for Realness}
\label{sec:realness}

First we discuss our results regarding the realness of our data; i.e., our results regarding A1. 
To do so, we use the \textsc{Classify} algorithm and the problems \textsc{Real}, \textsc{SynthRegular} and \textsc{SynthOutlier}. 
For realness, we will not use the \textsc{Synth} problem, which is fully synthetic data. 
The reason is that we are interested in the realistic reconstruction of regular instances. 
Whether these remain realistic in combination with the different types of generated outliers is less important. However, how the synthetic outliers on their own impact classification performance is investigated with \textsc{SynthOutlier}.

\subsection{Synthesizing Regular Instances}
\label{sec:real_synthregular}

\begin{figure}[ht]
	\centering
	\subfloat[Difference in Kappa.]{%
		\includegraphics[width=0.48\linewidth]{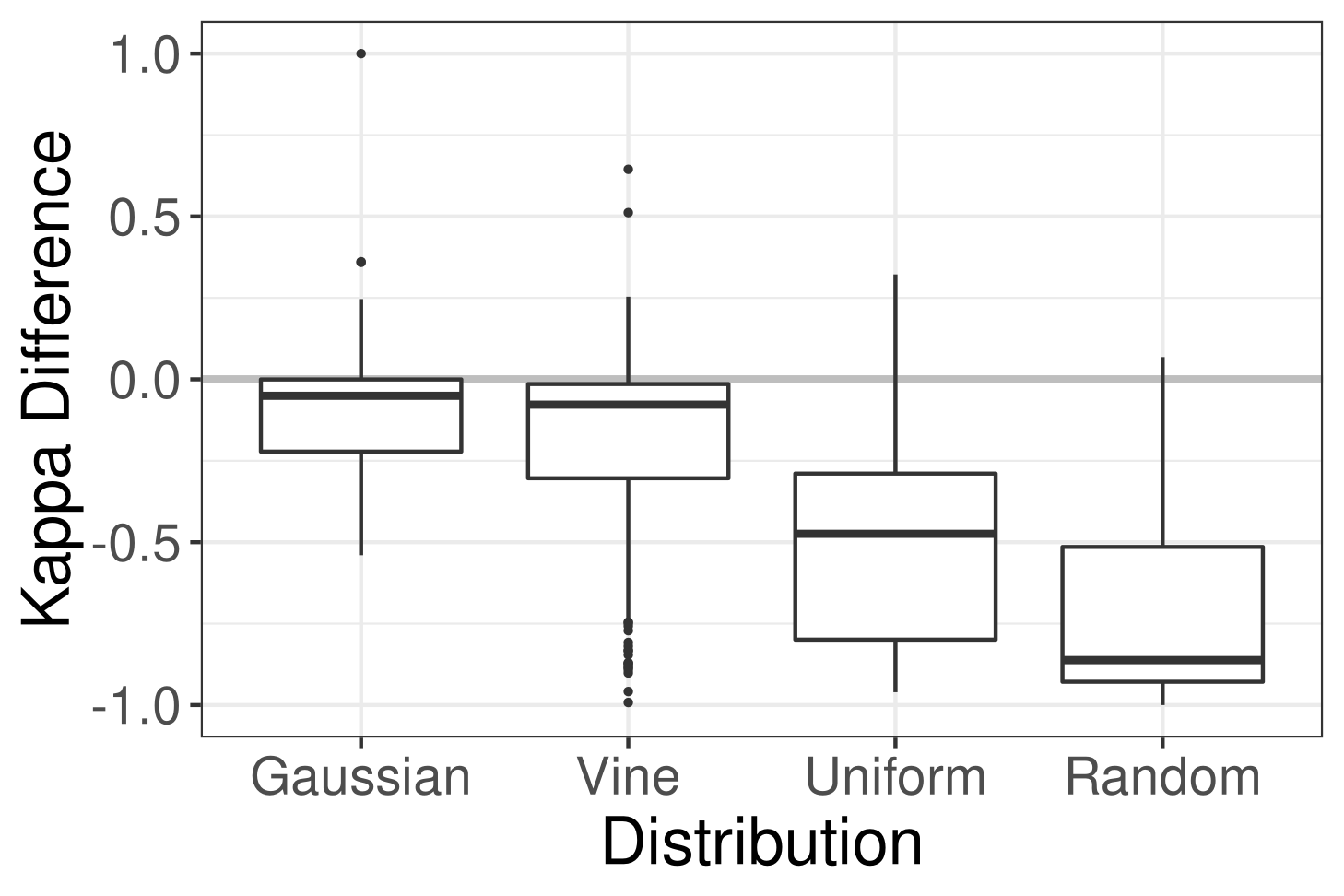}
		\label{fig:classify-synthregular-overall}%
		\Description{Boxplots with the difference in Kappa when training a classifier with synthetic regular instances, and a boxplot with the respective difference when using random guessing. The difference with Gaussian mixtures and the vine copula are small (i.e., close to zero). For the uniform distribution the median is about -0.5, but still much higher than the median from random guessing.}
	}
	\hfil
	\subfloat[Raw Kappa values.]{%
		\includegraphics[width=0.48\linewidth]{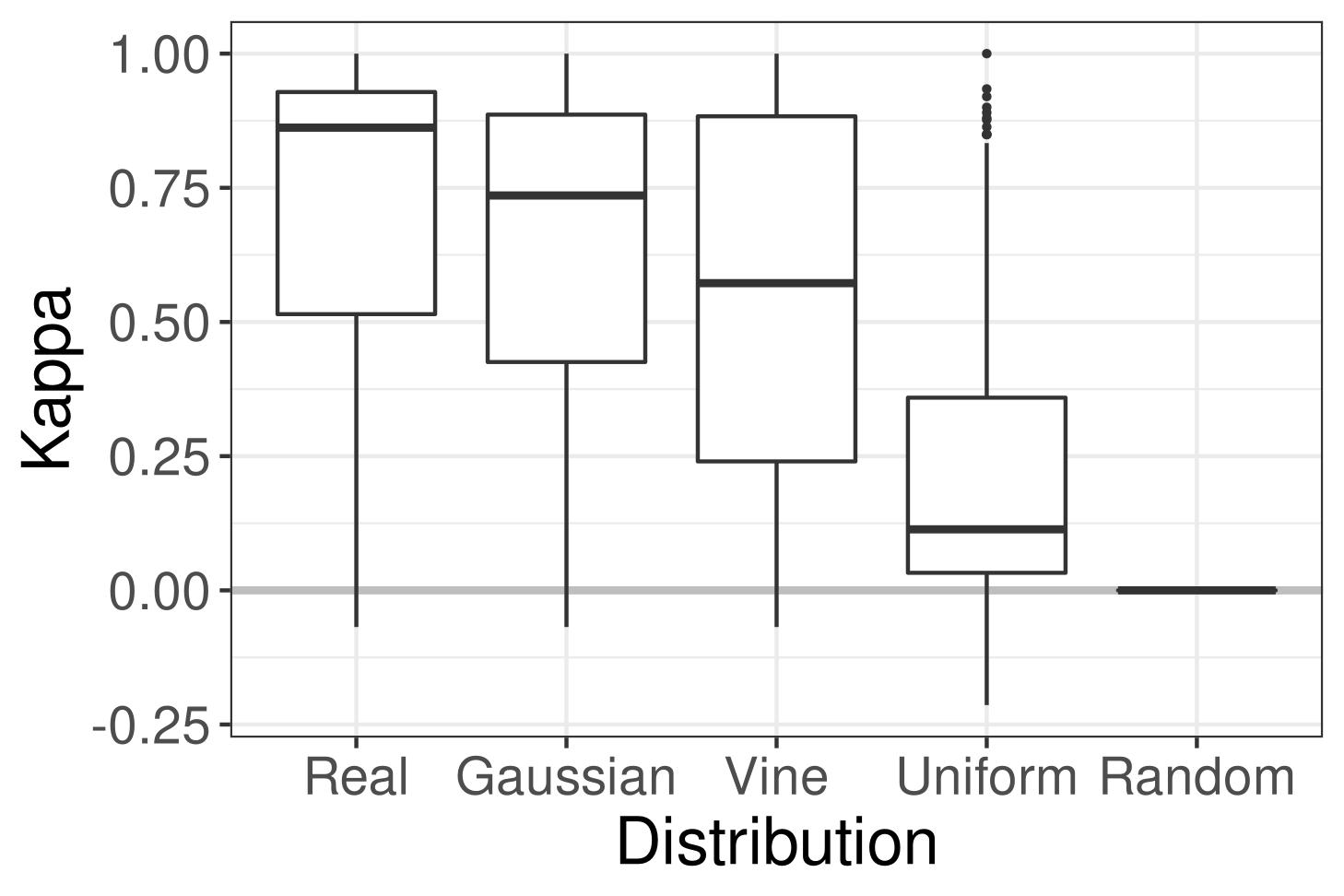}
		\label{fig:classify-synthregular-raw}%
		\Description{Boxplots with the raw Kappa when training a classifier with synthetic regular instances, the genuine regular instances, and when using random guessing. The boxes when regular instances are generated with Gaussian mixtures and the vine copula are close to the box from genuine regular instances. The box from the uniform distribution is close to zero, which is always the value for random guessing.}
	}
	\caption{Kappa with synthetic regular instances.}
	\label{fig:reduction_kappa_regulars}
\end{figure}

\cref{fig:reduction_kappa_regulars} graphs the realness of our synthetically reconstructed regular instances.
While \cref{fig:classify-synthregular-overall} displays the differences in Kappa when using synthetic regular instances, \cref{fig:classify-synthregular-raw} shows the raw Kappa values with real or synthetic regular instances.
The x-axis lists the distributions that are fitted to regular instances.
\emph{Random} is not connected to a distribution but shows the drop in classification performance when randomly guessing the class of instances. 
It is included as a reference point. 
The y-axis shows the drop in Kappa when the classifier is trained with synthetic regular instances instead of real ones. 

From \cref{fig:reduction_kappa_regulars} we can see that synthetic regular instances from every distribution yield a better classifier than random guessing --- even from the uniform distribution. 
Next, the Gaussian mixtures give the best reconstruction overall but the vine is not far from it. 
Thus, Gaussian mixtures yield the best reconstruction of regular instances.
Since the median is close to zero, we conclude that our reconstructed regular instances are \emph{realistic}.

\subsection{Synthesizing Outliers}

\begin{figure}[ht]
	\centering
	\subfloat[Outliers from a characterizable deviation.]{%
		\includegraphics[width=0.48\linewidth]{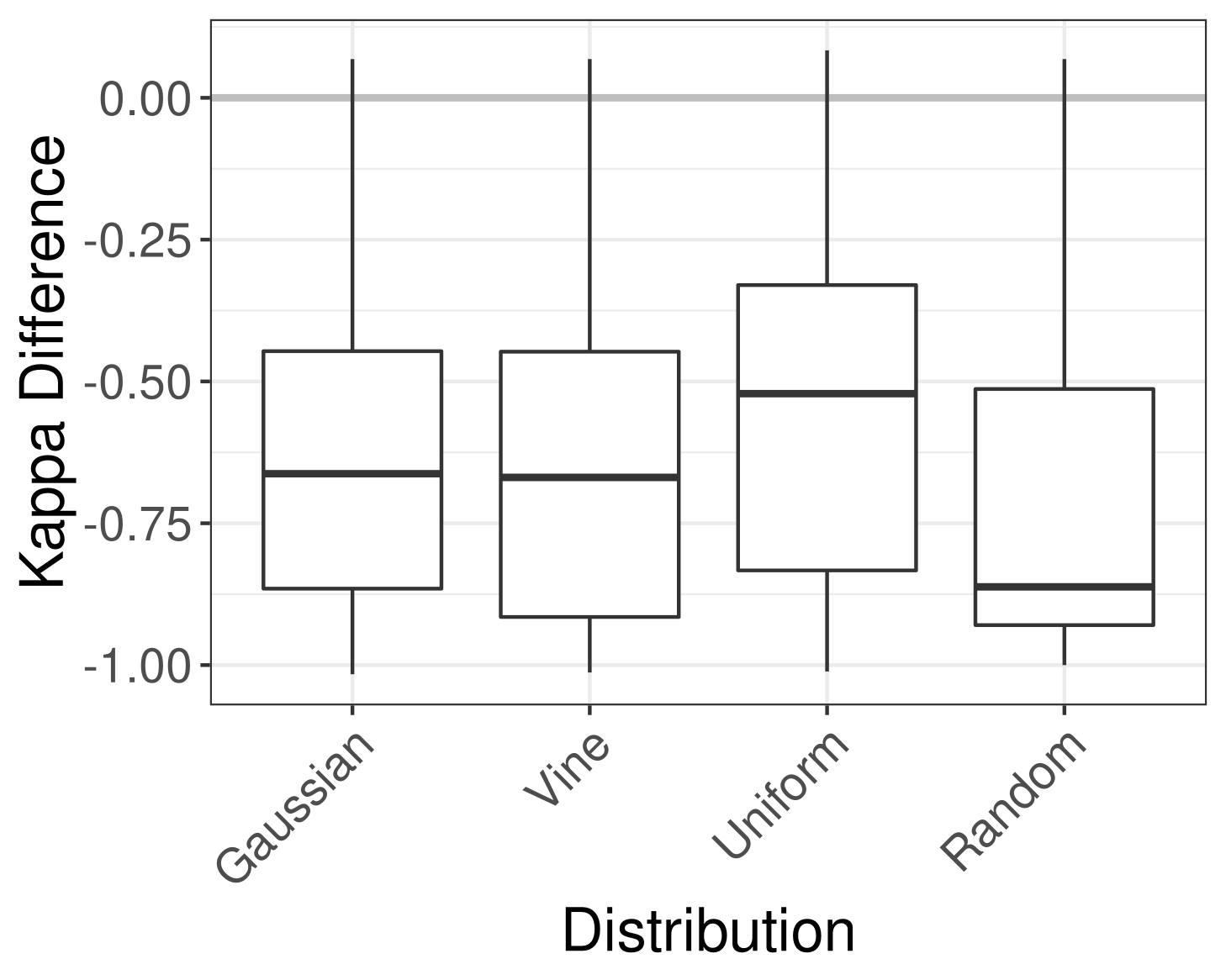}
		\label{fig:classify-synthoutlier-chracterizable}%
		\Description{Boxplots with the difference in Kappa when training a classifier with characterizable synthetic outliers, and a boxplot with the respective difference when using random guessing. The difference with all distributions are highly negative. For the uniform distribution the median is the highest with about -0.5.}
	}
	\hfil
	\subfloat[Outiers like the genuine ones.]{%
		\includegraphics[width=0.48\linewidth]{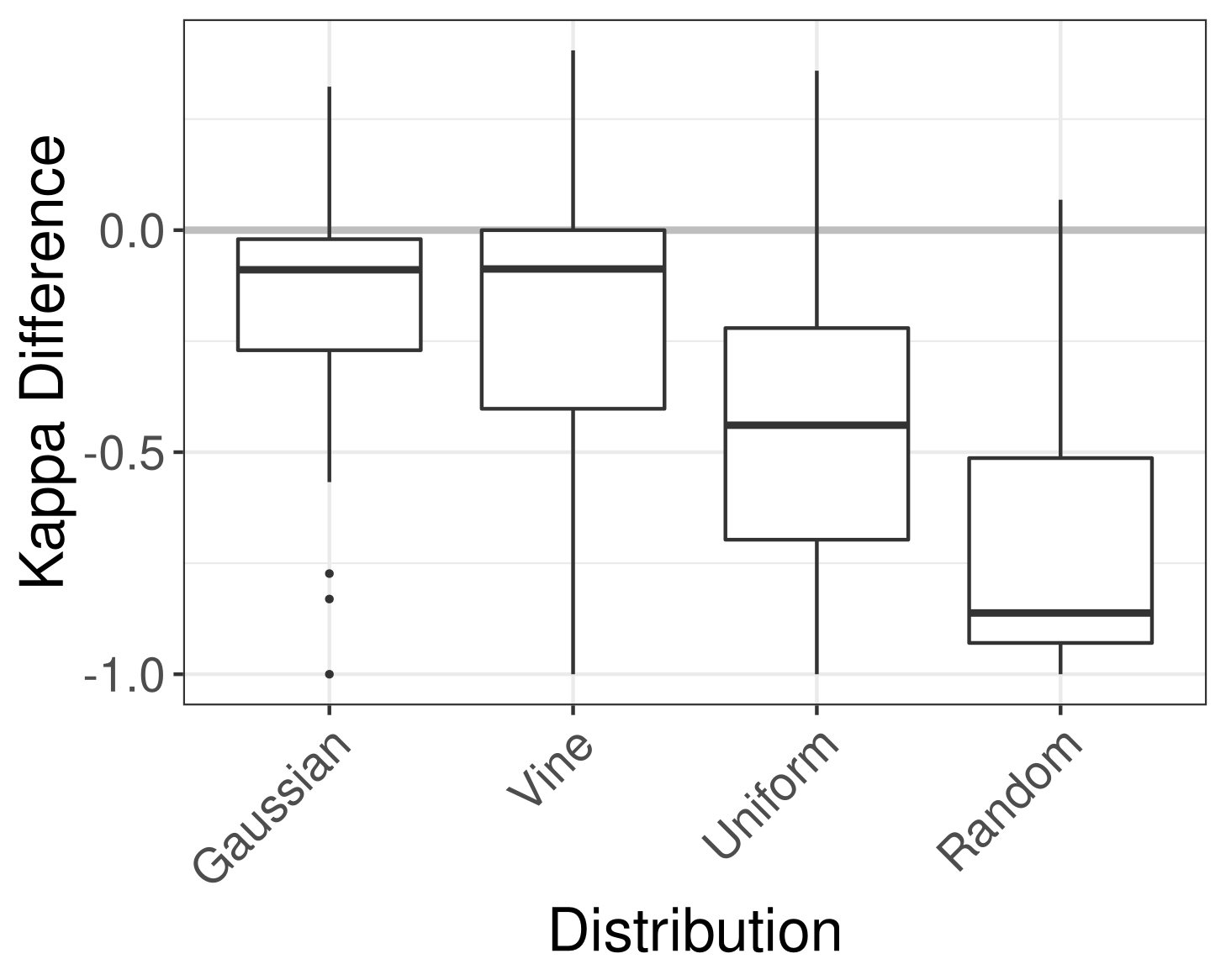}
		\label{fig:classify-synthoutlier-genuine_outs}%
		\Description{Boxplots with the difference in Kappa when training a classifier with synthetic outliers reconstructing genuine outliers, and a boxplot with the respective difference when using random guessing. The difference with Gaussian mixtures and the vine copula are small (i.e., close to zero). For the uniform distribution the median is about -0.5, but still much higher than the median from random guessing.}
	}
	\caption{Reduction in Kappa for synthetic outliers.}\label{fig:reduction_kappa_outs}
\end{figure}

Outliers should be credible and insightful deviations from the regular instances, i.e., should be characterizable.
It is only for completeness that we also investigate the drop in classification performance when training with synthetic outliers. 
See \cref{fig:reduction_kappa_outs}. 
Results in \cref{fig:classify-synthoutlier-chracterizable} are from characterizable synthetic outliers, as introduced in \cref{sec:instantiations}.
\cref{fig:classify-synthoutlier-genuine_outs} contains the results when the distribution of the synthetic outliers is derived from the real outliers and not a characterizable deviation from the regular instances. 
Results with this type of synthetic outliers serve as reference points.

There is a rather big drop in classification performance in \cref{fig:classify-synthoutlier-chracterizable}. 
It is expected since the characterizable synthetic outliers do differ from the real outliers in the data sets.
Observe that this is not a negative result since our process does not aim at the generation of outliers similar to the real ones.
Interestingly the uniform distribution results in the smallest drop in Kappa. 
We hypothesize that this is because of the one-class nature of the respective resulting classifier. 
\citet{steinwart_classication_2005} discuss this for an SVM classifier, i.e., with outliers generated from a uniform distribution the classifier might become a well calibrated one-class classifier. 
This would explain its good detection performance regarding real outliers.

\cref{fig:classify-synthoutlier-genuine_outs} is very similar to \cref{fig:classify-synthregular-overall}. 
I.e., Gaussian Mixtures and Vine result in only a small drop in Kappa overall. 
However, compared to regular instances, the Vine seems to perform slightly better for outliers. 
We find this particularly interesting in two regards. 
(1) It seems to indicate that the outliers in the benchmark data are not of extremely difficult nature. 
Common statistical distributions fit them reasonably well. 
(2) The uniform distribution is not a very good fit for the real outliers in the benchmark data. 
We see this as further evidence that uniform outliers do not represent real outliers well.

\section{Outlier Detection Benchmark}
\label{sec:benchmark}

In this section, we showcase (A2) and validate (A3) our process.
We perform a benchmark and compare its results to the ones obtained with fully real data.

In our benchmark, we assess detection performance with the area under the precision recall curve (AUC PR).
With imbalance, the precision-recall curve is preferable over the ROC curve \citep{saito_precision-recall_2015}; so we do not display our results using the area under the ROC curve (AUC ROC) in this article. 
However, since the AUC ROC is so widely known, we do provide our results in terms of AUC ROC together with our code.

The most trivial baseline for outlier detection is random guessing whether an instance is an outlier. 
The AUC PR with such random guessing depends on the percentage of actual outliers in the data set.
While this percentage is always 5\% in our synthetic data, in the real data it might be less. 
The reason for the possibly smaller percentage with real data is that some real-world data sets have less than 5\% labeled outliers.
Since we are not aware of any gold standard to increase the number of outliers with real data, sticking to the available outliers in these cases appears to be most meaningful.
For fair comparisons, we adjust the AUC PR so that random guessing always yields 0. 

Next we compare results from our synthetic data to ones from real-world data. Then we analyze the correlation of the ideal scorings and the scores from the different outlier detection methods.

\subsection{Comparison to Real Data}

\begin{figure*}[ht]
	\centering
	\includegraphics[width=1\linewidth]{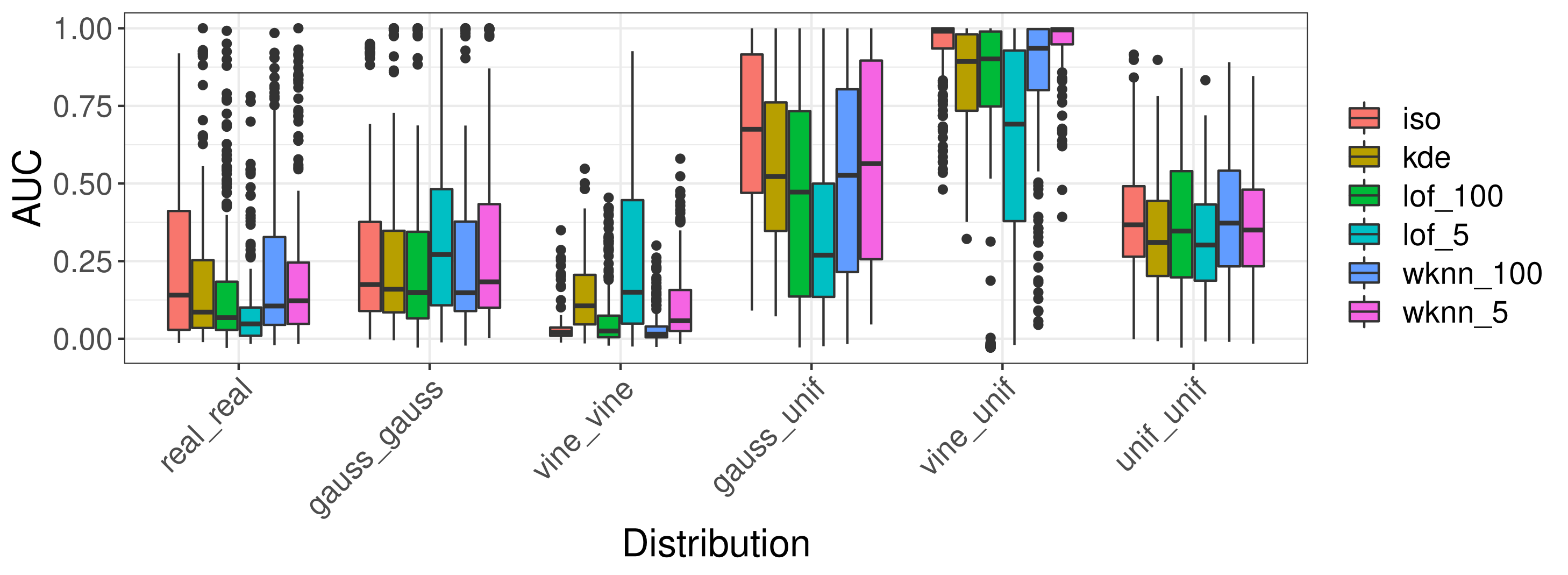}
	\caption{Performance of the different detection methods on synthetic and real data.}
	\Description{Boxplots with the AUC PR of different outlier detection methods with our different instantiations and fully real data. With local and dependency outliers a very local detection method performance best, while global methods perform better in the other cases.}
	\label{fig:detect-bestmethod}
\end{figure*}

\cref{fig:detect-bestmethod} contains results regarding our benchmark for different instantiations and the real data. The y-axis gives the respective AUC PR, and
the x-axis gives the distribution regular instances (term before "\_")  or outliers (term after "\_") are samples from.
The results with fully real data suggest that the Isolation Forest or w$k$NN approach perform well overall --- just as observed with previous benchmarks. 
Detection approaches rank similarly with synthetic data that uses the uniform distribution to generate outliers (gauss\_unif, vine\_unif and unif\_unif). 
The local method "lof\_5" performs worst overall for all three types of data with global outliers. 
This is expected.

The two types of synthetic data that do not use the uniform distribution give a different ranking. 
Both suggest the local method "lof\_5" to be best overall. 
For the synthetic data with outliers from a Gaussian mixture this is what we would expect. The outliers are designed to be local ones. 
Nevertheless, this nicely confirms that this type of synthetic data is well suited to evaluate methods for local outlier detection. 
With outliers in the dependency structure of the data (vine\_vine) the isolation Forest, "lof\_100" and "wknn\_100" do not perform well overall. 
Their AUC PR is close to 0. Hence, local approaches seem to detect such outliers much better.

Regarding the overall AUC PR, it is the highest for synthetic data with uniform outliers. 
I.e., every approach has a higher average AUC PR than with real data. 
This is close to what \citet{emmott_meta-analysis_2015} found with their simple synthetic data. 
With the other two forms of synthetic data, this is not the case --- regarding the vine in particular. 
However, in our opinion the bare value of the AUC PR is not very meaningful for synthetic data. 
It depends directly on $\alpha$ from the instantiation based on Gaussian Mixtures for example, and it can be adjusted accordingly.
Instead, the ranking in terms of different characteristics is more conclusive.

\subsection{Ideal Scores}

\cref{sec:ideal_scores} has introduced three ideal scorings. Here we analyze our benchmark results with regard to them. 
First we investigate the detection performance (AUC PR) of our ideal scorings.
Then we study the relationship of the ranking of instances from the detection methods to the ones from the ideal scorings.

\subsubsection{Detection Performance}

\begin{figure}[ht]
	\centering
	\subfloat[Detection Quality.]{%
		\includegraphics[width=.385\linewidth]{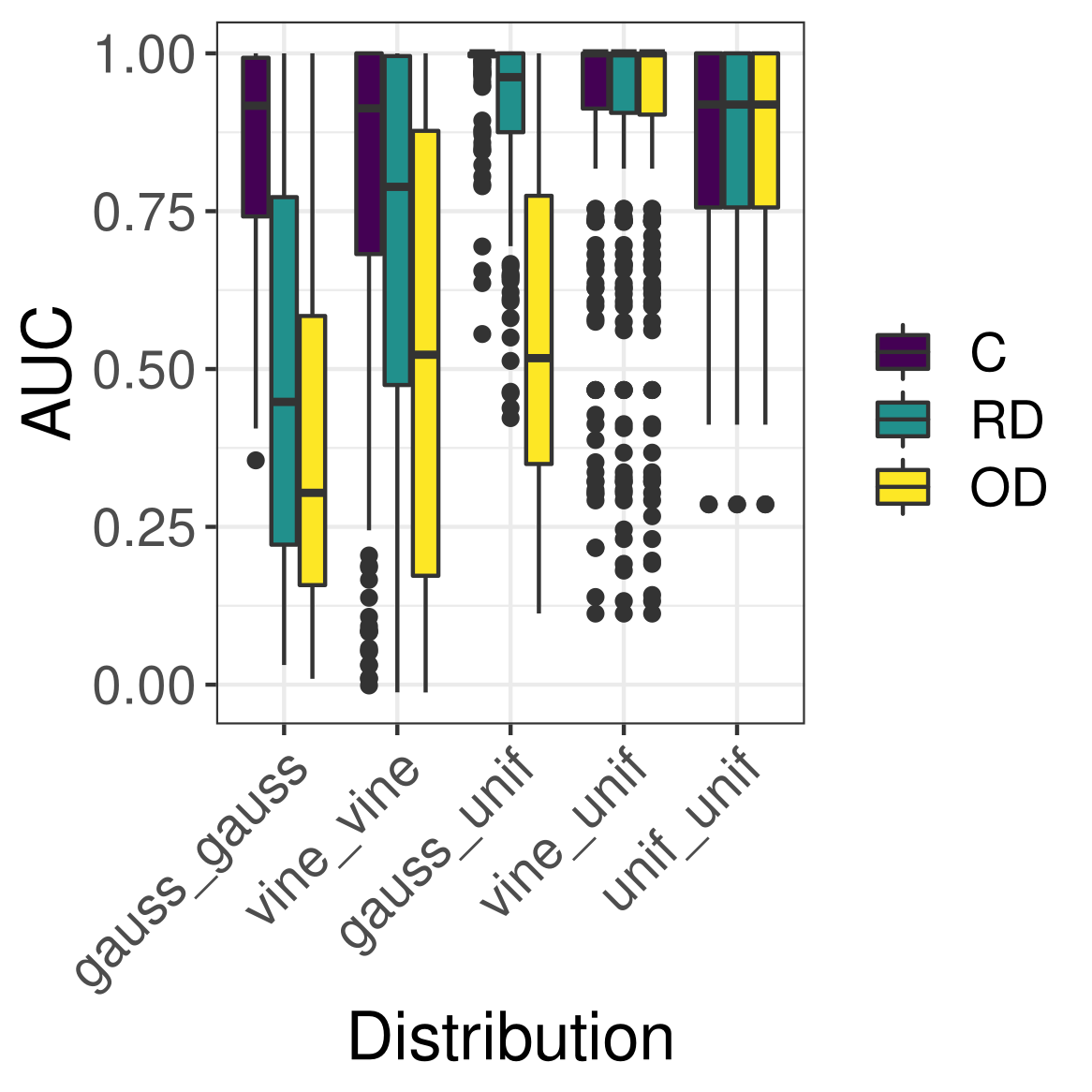}
		\label{fig:detect-ideal-overall}%
		\Description{Boxplots with the AUC PR of our ideal scores with our different instantiations.}
	}
	\subfloat[Correlation (Colors from \cref{fig:detect-bestmethod}).]{%
		\includegraphics[width=.575\linewidth]{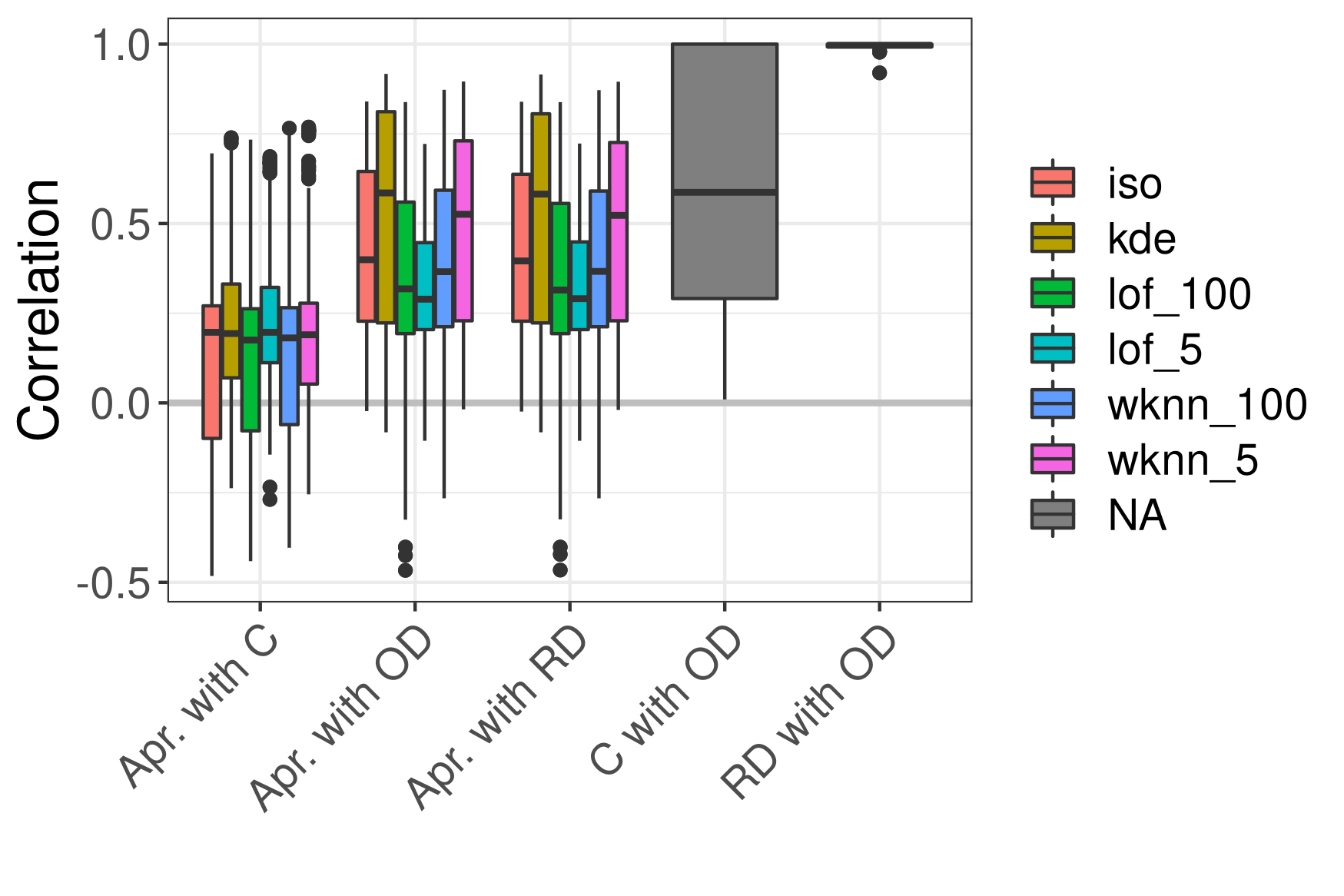}
		\label{fig:detect-correlation}%
		\Description{Boxplots with the Correlation of our ideal scores with themselves but also the scores from the detection methods.}
	}
	\caption{Ideal scores.}
	\label{fig:ideal}
\end{figure}

\cref{fig:detect-ideal-overall} displays the detection quality of each of the ideal scorings. 
The ranking reflects our expectations: C has the highest AUC PR, followed by RD and then OD. 
I.e., the fully supervised case (C) yields better results than the case in which the distribution of outliers is not modeled (RD) or both classes are modeled together (OD). 
When regular instances and outliers are from a uniform distribution all ideal scores are the same. 
The reason is that in this case the densities $\densout$ and $\densreg$ only differ by a constant. 
Since the AUC PR is computed by varying the threshold on the scorings, the specific value for the threshold does not affect it. 
So all ideal scores give the same AUC PR. 
When outliers come from a uniform distribution, the AUC PR is highest overall. 
This confirms their rather global nature.

\subsubsection{Relationship to Detection Methods}

 To analyze how well the scorings from the detection methods coincide with our ideal ones, we use the Kendall Tau \citep{kendall_new_nodate}. 
 It quantifies the similarity of the rankings with the two scorings. 
 See \cref{fig:detect-correlation}. The scores of the outlier detection approaches are abbreviated with "A".

The two ideal scorings based on density (RD and OD) have an extremely strong positive correlation. Hence, using the density of all instances as a proxy for the one of regular instances
does not impact the overall ranking of instances by much. 
We find this very interesting: 
In the unsupervised setting, computing the overall density is simple -- no labels are needed. 
Getting an estimate of the density of regular instances is much more challenging.
However, it seems that the small difference in the overall ranking does affect outlier detection performance. 
This performance varies substantially for the different rankings (cf.\ \cref{fig:detect-ideal-overall}). 
The classification scoring does not correlate much with the low density scorings. 
This is expected though: The classify scoring uses the actual distributions of outliers and regular instances. 
Thus, it can also distinguish instances that might not be meaningful outliers. 
An example is generated outliers that are extremely close to regular instances. 
The approaches in general correlate much more with the density scoring than with the classify scoring.
KDE in particular has a strong correlation with density --- which is expected. 
However, since it is not the best outlier detector (cf.\ \cref{fig:detect-bestmethod}), there seems to be more to outlier detection than just density estimation. 
While "lof\_5" does not correlate with density as well as the other approaches, it does correlate with the classify scoring more than some of the other approaches.
We hypothesize that this comes from its great detection capabilities with the Gaussian Mixture but also vine outliers.

\section{Conclusions}
\label{sec:conclusion}

In this section we first we summarize our work, then discuss its limitations, and afterwards give some possible directions for future work.

\subsection{Summary}

Benchmarking unsupervised outlier detection methods continues to be difficult.  
We have shown that the problem can be dealt with using fully synthetic data, where outliers are chracterizable deviations from regular instances. 
In this work we have proposed a process that yields such synthetic data. 
Using existing real-world data sets as basis for the generation renders the generated data realistic. 
The combination of realistic data and outliers as characterizable deviations allows for high interpretability.
We also propose concrete instantiations of our process for three types of outliers, each one exhibiting different characteristics. 
An extensive benchmark with state-of-the-art unsupervised outlier detection methods confirms the usefulness of our proposed process. 
Our results suggest that the relative performance of the detection methods does indeed differ with the various types of outliers. 
Put differently, no method is optimal for all types.
The synthetic nature of our data also allowed for the computation of some ideal scorings. 
These ideal scorings give way to further interpretations, for example how far the detection performance of certain methods is from the performance of such an ideal scoring.

\subsection{Limitations}

While the process proposed here is a significant advancement in the area of outlier detection benchmarks, it is just a first stab at the problem and has some limitations, as follows.

\subsubsection{Favoring Certain Detection Methods}

All three of our instantiations rely on well-known statistical distributions for the generative models.
However, statistical distributions can be used for outlier detection themselves.
For example, \citet{ren_fault_2017} make use of a vine copula for outlier detection.
Clearly, comparing other detection methods to one based on a vine copula can be biased when the data is generated from a vine copula.
However, this bias can be expected. 
So one could eliminate this bias by not including such methods in the benchmark.
In this regard we hypothesize that with statistical distributions theoretical analyses might be more conclusive than experiments.

\subsubsection{Realness is Measured with a Classifier}

In our experiments we measure the realness of our data using a classifier. 
If the classification performance of a classifier trained with synthetic instances instead of real ones is similar to a classifier trained with real instances, the synthetic data is realistic (we argue). 
However, the degree of this realness depends on the classifiability of a specific data set:
If a classifier has a rather low classification accuracy on real data, a similarly low accuracy with synthetic data is not surprising. 
On the other hand, measuring realness by means of classification is practical. 
It is easily applicable to many different data sets, as opposed to verification by domain experts.
Additionally, the real data sets used in this study yield a rather high classification accuracy overall.
So this issue should have only a small effect on our results.

\subsubsection{No Outliers in Labeled Regular Instances}

Another limitation in terms of realness could be that we assume the instances in the real-world data to be labeled correctly as regular.
Put differently, we assume that the labels of regular instances in common benchmark data are reliable.
This is not necessarily the case, as we can already see in \cref{exa:motivation}.
In the regular instances from the Wilt data set there is a clear outlier.
Incorrect labels of regular instances do not affect the characteristics of the generated outliers -- with respect to the regular instances, the characteristics remain as intended.
However, it possibly affects the realness of the generated regular instances.
In other words, this might lead to the generation of regular instances in regions where no regular instances should lie from a domain perspective.
A possible solution could be to first eliminate any inlier that might be an outlier. 
For example, one could do so by applying a very conservative outlier detection method before fitting a model to the inliers. 
However, this conservative detection method would then be another parameter of our process. 
Studying the effect of such a parameter would exceed the scope of this article.

\subsubsection{Infinite Number of Outlier Types}
\label{sec:infinite_types_outliers}

This article does not feature every possible type of outlier. 
There are infinitely many types of outliers conceivable. 
To illustrate, think of the dependency outliers: 
With the specific instantiation we propose for such outliers, dependency outliers are characterized by not following any dependency between attributes. 
Thus, with a dependency outlier from our respective instantiation, the value of one attribute is independent of the values in any other attribute.
However, a slightly different type of dependency outlier could feature a certain dependency between attributes. 
To qualify as outliers, this dependency must differ substantially from the one that regular instances exhibit. 
Since there is an infinite number of ways attributes could depend on each other, there already is an infinite number of possibilities for this type.
Our focus in this work has been on types that are common with existing detection methods (e.g., local ones from the Local Outlier Factor (LOF)) and that are feasible in many different domains. 

\subsection{Future Work}

A promising future direction might be to develop approaches for more specialised types of outliers.
One example are dependency outliers with a specific dependency structure, as just mentioned.
Statistical distributions that describe categorical data are interesting in this regard as well.
Another direction is investigating the properties of outliers and of regular instances that affect the parameterization of outlier detection methods. 
Choosing these values still is a challenge \citep{campos_evaluation_2016}, and our proposed process might be of great help here. 
Investigating the gap between the performance of the detection methods and our ideal scorings might also be a useful indicator for improved future detection methods.
All in all, our proposed process allows for a realistic comparison of detection methods and for meaningful interpretations of results.

\begin{acks}
	This work has been supported by the \grantsponsor{dfg}{German Research Foundation (DFG)}{} as part of the \grantnum{dfg}{Research Training Group GRK 2153: Energy Status Data -- Informatics Methods for its Collection, Analysis and Exploitation}. 
	We also thank Lena Witterauf for much support.
\end{acks}


\bibliographystyle{ACM-Reference-Format}
\bibliography{benchmarks,synth-data,others,artouts}

\end{document}

%% file: comands.tex
\newcommand\realData{\textit{realData}}
\newcommand\synthData{\textit{synthData}}
\newcommand\R{{\rm I\!R}}
\newcommand\nDim{d}
\newcommand\nReal{n}
\newcommand{\instance}{\textit{inst}}

\newcommand\model{\mathcal{M}}
\newcommand\mReg{\model_{\textit{Reg}}}
\newcommand\mOut{\model_{\textit{Out}}}
\newcommand{\densreg}{\densNoArg_{\textit{Reg}}}
\newcommand{\densout}{\densNoArg_{\textit{Out}}}

\newcommand{\funArg}[1]{
	\ifthenelse{\isempty{#1}}{
		\cdot
	}{
		#1
	}
}
\newcommand{\dectNoArg}{\text{dect}}
\newcommand{\dect}[1]{\dectNoArg\!\left(\funArg{#1}\right)}
\newcommand{\fitNoArg}{\text{fit}}
\newcommand{\fit}[1]{\fitNoArg\!\left(\funArg{#1}\right)}
\newcommand{\genNoArg}{\text{gen}}
\newcommand{\gen}[1]{\genNoArg\!\left(\funArg{#1}\right)}
\newcommand{\densNoArg}{\text{dens}}
\newcommand{\dens}[2]{\densNoArg\!\left(\funArg{#1, #2}\right)}
\newcommand{\modifyNoArg}{\text{modify}}
\newcommand{\modify}[1]{\modifyNoArg\!\left(\funArg{#1}\right)}